%% file: MREPlanner_ICRA_2022.tex
\documentclass[letterpaper, 10 pt, conference]{ieeeconf}  

\IEEEoverridecommandlockouts                              

\usepackage{graphics} 
\usepackage{graphicx}

\usepackage{epsfig} 
\usepackage{amsmath} 
\usepackage{amssymb}  

\usepackage{algorithm}
\usepackage[noend]{algpseudocode} 
\algnewcommand\AAND{\textbf{ and }}
\algnewcommand\Or{\textbf{ or }}

\usepackage{color}
\usepackage{citesort}
\usepackage{xurl}
\usepackage[breaklinks]{hyperref}

\DeclareMathAlphabet{\pazocal}{OMS}{zplm}{m}{n}

\newcommand{\Bs}{\pazocal{B}}

\newcommand{\Ss}{\pazocal{S}}

\newcommand{\Is}{\pazocal{I}}

\newcommand{\Ms}{\pazocal{M}}

\newtheorem{definition}{Definition}

\DeclareMathAlphabet{\mathpzc}{OT1}{pzc}{m}{it}

\usepackage{array}
\newcolumntype{C}[1]{>{\centering\arraybackslash}p{#1}}
\newcolumntype{M}[1]{>{\raggedright\arraybackslash}p{#1}}

\usepackage{array} 
\newcolumntype{L}[1]{>{\raggedright\let\newline\\\arraybackslash\hspace{0pt}}m{#1}}	
\newcolumntype{S}[1]{>{\centering\let\newline\\\arraybackslash\hspace{0pt}}m{#1}}
\newcolumntype{R}[1]{>{\raggedleft\let\newline\\\arraybackslash\hspace{0pt}}m{#1}}

\newtheorem{problem}{Problem}

\makeatletter
\renewcommand*{\@opargbegintheorem}[3]{\trivlist
  \item[\hskip \labelsep{\itshape #1\ #2}] \textit{(#3)}\ }
\makeatother

\title{\LARGE \bf
Autonomous Teamed Exploration of Subterranean Environments using Legged and Aerial Robots
}

\author{Mihir Kulkarni$^{1,3,\star}$, Mihir Dharmadhikari$^{1,3,\star}$, Marco Tranzatto$^2$, Samuel Zimmermann$^2$, Victor Reijgwart$^2$, \\ Paolo De Petris$^3$, Huan Nguyen$^{3}$, Nikhil Khedekar$^{3}$, Christos Papachristos$^{1}$,\\ Lionel Ott$^{2}$, Roland Siegwart$^{2}$, Marco Hutter$^{2}$, and Kostas Alexis$^{3}$
\thanks{This material is based upon work supported by the Defense Advanced Research Projects Agency (DARPA) under Agreement No. HR00111820045. The presented content and ideas are solely those of the authors.}
\thanks{$^{1}$University of Nevada, Reno, 1664 N. Virginia, 89557, Reno, NV, USA
        {\tt\small mkulkarni@nevada.unr.edu}}
\thanks{$^2$ETH Zurich, Leonhardstrasse 21, 8092, Zurich, Switzerland}%
\thanks{$^{3}$NTNU, O. S. Bragstads Plass 2D, 7034, Trondheim, Norway}
\thanks{$^{\star}$ The authors have contributed equally.}
}

\begin{document}

\maketitle
\thispagestyle{empty}
\pagestyle{empty}

\begin{abstract}

This paper presents a novel strategy for autonomous teamed exploration of subterranean environments using legged and aerial robots. Tailored to the fact that subterranean settings, such as cave networks and underground mines, often involve complex, large-scale and multi-branched topologies, while wireless communication within them can be particularly challenging, this work is structured around the synergy of an onboard exploration path planner that allows for resilient long-term autonomy, and a multi-robot coordination framework. The onboard path planner is unified across legged and flying robots and enables navigation in environments with steep slopes, and diverse geometries. When a communication link is available, each robot of the team shares submaps to a centralized location where a multi-robot coordination framework identifies global frontiers of the exploration space to inform each system about where it should re-position to best continue its mission. The strategy is verified through a field deployment inside an underground mine in Switzerland using a legged and a flying robot collectively exploring for $45\textrm{ min}$, as well as a longer simulation study with three systems. 

\end{abstract}

\section{INTRODUCTION}\label{sec:intro}
\input{Introduction.tex}

\section{RELATED WORK}\label{sec:related}
\input{RelatedWork.tex}

\section{PROBLEM FORMULATION}\label{sec:probstat}
\input{ProblemFormulation.tex}

\section{PROPOSED APPROACH}\label{sec:approach}
\input{Approach.tex}

\section{EVALUATION STUDIES}\label{sec:evaluation}
\input{EvaluationStudies.tex}

\section{CONCLUSIONS}\label{sec:concl}
\input{Conclusion.tex}

\bibliographystyle{IEEEtran}
\bibliography{./MREPlanner_ICRA_2022}

\end{document}

%% file: Introduction.tex
The collective progress in robotic systems has enabled their utilization in a multitude of autonomous exploration and mapping missions. Aerial and ground robots are now employed in diverse search and rescue~\cite{balta2017integrated,tomic2012toward,michael2012collaborative,delmerico2019jfr}, industrial inspection~\cite{SIP_AURO_2015,BABOOMS_ICRA_15,caprari2012highly,balaguer2000climbing,sawada1991mobile,gehring2019anymal,hutter2018towards,Kolvenbach2019fsr}, surveillance~\cite{grocholsky2006cooperative}, planetary exploration~\cite{schilling1996mobile,bares1989ambler,schuster2019towards,welch2013systems,balaram2021ingenuity,boeder2020mars} and other mission scenarios. Despite the progress, a set of environments remain particularly demanding for robots to autonomously explore and challenge the state-of-the-art in robot navigation and autonomy. Among them, subterranean settings such as underground mines, subway infrastructure and cave networks are especially strenuous. Motivated by the goals and aspirations of the DARPA Subterranean Challenge~\cite{darpasubt}, this work culminates on a sequence of developments in autonomous exploration path planning for both legged and aerial robots that have enabled the robots of Team CERBERUS to explore underground mines, multi-level power plants, tunnels, sewers, caverns, lava tubes and more. 

%
\begin{figure}[h!]
\centering
    \includegraphics[width=0.99\columnwidth]{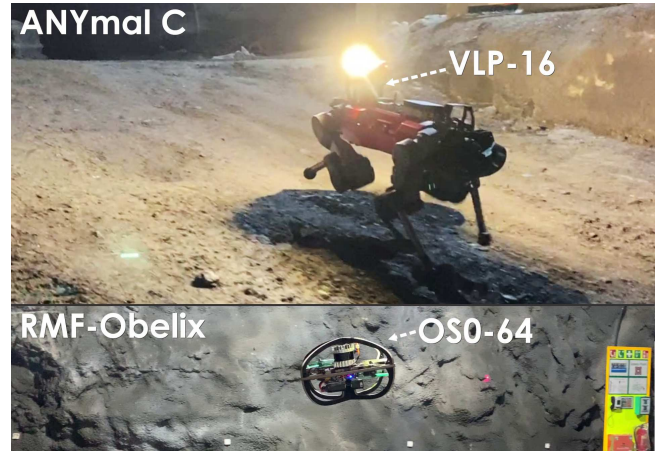}
\vspace{-4ex}
\caption{Instances of autonomous exploration of a subterranean environment.}\label{fig:motivation}
\vspace{-5ex}
\end{figure}
%

Specifically, we present two synergistic contributions, namely a) ``GBPlanner2'', a revised and enhanced version of our Graph-Based subterranean exploration path Planner~\cite{GBPLANNER_JFR_2020} to uniformly guide both legged and aerial robots (open-sourced at \url    {https://s.ntnu.no/gbplanner}), as well as b) the newly developed CoOperation for HeterOgeneous Robot Teams (COHORT) framework that facilitates the synergistic exploration of large-scale, multi-branching, and diverse subterranean settings having both narrow and wide sections. The two methods working in harmony enable the autonomous cooperative exploration and mapping of complex underground domains using legged and flying systems that coordinate their exploration, while each retains complete individual exploratory capacity.

To verify the proposed solution for teamed exploration path planning, a set of evaluations are presented including a deployment in the Hagerbach underground mine in Switzerland (Figure~\ref{fig:motivation}) using legged and aerial robots exploring for $37~\textrm{min}$ and $8~\textrm{min}$ respectively, as well as a large-scale simulation study. We demonstrate that due to the individual properties of GBPlanner2 and COHORT, as well as their synergy, the robots present efficient autonomous exploration behaviors that are resilient against the complexities introduced by the vast scale of the environments, the challenging terrain and complicated geometries involved, as well as the difficulty of establishing reliable communications. A page of results is maintained at \url{https://s.ntnu.no/exploration}.


In the remaining paper, Section~\ref{sec:related} presents related work, followed by the problem statement in~\ref{sec:probstat}. The proposed approach is detailed in Section~\ref{sec:approach}, with evaluation studies in Section~\ref{sec:evaluation} and conclusions in Section~\ref{sec:concl}.

%% file: RelatedWork.tex
A rich body of work has focused on the problems of autonomous single and multi-robot exploration~\cite{RHEM_ICRA_2017,NBVP_ICRA_16,yoder2016autonomous,connolly1985determination,yamauchi1997frontier,GBPLANNER_JFR_2020,MBPlanner_ICRA_2019,miller2020mine,pei2010coordinated,burgard2005coordinated,cesare2015multi,corah2021volumetric,goel2021fast,goel2020rapid,smith2018distributed}. Early work in single-robot exploration included the sampling of ``next--best--views''~\cite{connolly1985determination}, and the detection of frontiers~\cite{yamauchi1997frontier}, while recent efforts have focused on powerful planning techniques such as random trees and graphs possibly combined with volumetric calculations~\cite{NBVP_ICRA_16,VSEP_ICRA_2018,RHEM_ICRA_2017,HABPlanner_ICRA_2021,yang2021graph,lee2021real}, receding horizon techniques~\cite{NBVP_ICRA_16}, multi-objective optimization~\cite{VSEP_ICRA_2018,HABPlanner_ICRA_2021}, information-theoretic schemes~\cite{tabib2016computationally}, learning-based methods~\cite{LBPlanner}, and approaches that account for the likelihood of accumulating localization drift~\cite{RHEM_ICRA_2017,schmid2021unified}. In multi-robot exploration, the seminal work in~\cite{burgard2005coordinated} presented a strategy for multi-robot coordination exploiting a grid map and a planning policy that tries to minimize the collective exploration time by considering both the cost of reaching a certain frontier cell and the ``exploration utility'' of each such cell as a function of the number of robots moving to that cell. Recent efforts have included distributed inference-based schemes~\cite{smith2018distributed}, volumetric gain-driven multi-robot exploration~\cite{corah2021volumetric}, methods on the collaboration of ground and aerial systems~\cite{delmerico2017active}, techniques that consider the effect of limited communication range and endurance~\cite{cesare2015multi,pei2010coordinated}, concepts on how robots can work together to physically clear blocked paths~\cite{andre2016collaboration} and more. The community has recently focused on the specific problem of subterranean exploration given the accelerating effects of the DARPA Subterranean Challenge. In response, groups around the world have proposed novel methods both for single- and multi-robot exploration. This includes techniques for quadruped robots~\cite{miller2020mine}, methods tailored to fast exploration using aerial platforms~\cite{MBPlanner_ICRA_2019}, schemes for both legged and flying systems~\cite{GBPLANNER_JFR_2020}, hierarchical frameworks to exploit dense local and sparse global information~\cite{caotare}, multi-robot exploration strategies~\cite{goel2020rapid}, approaches exploiting underground mine and cave topologies~\cite{mansouri2020mav,petracek2021large}, and significant field robotics work~\cite{CERBERUS_SUBT_PHASE_I_II,agha2021nebula,rouvcek2019darpa,explorer_phase_i_ii}. Motivated by the importance of autonomous exploration and tailored to the teamed deployment of legged and flying robots inside subterranean settings, this work contributes two methods, namely on teamed exploration coordination and single-robot planning that enable resilient multi-robot teaming and reliable single-robot operation when communication to and from a robot is not possible, ability to negotiate challenging terrain and capacity to map diverse and large-scale geometries.

%% file: ProblemFormulation.tex
The overall problem considered in this work is that of autonomously exploring a bounded volume $V\subset\mathbb{R}^3$ enclosing a subterranean environment with a heterogeneous team of robots given models of their motion constraints, as well as of their onboard depth sensors $\{\mathbb{S}^i\}$ with horizontal and vertical Fields Of View (FoV) $[F_H^i,F_V^i]$, and effective range $d_{\max}^i$. As subterranean environments such as underground mines, subway infrastructure, and caves often consist of complex networks of branches, multiple levels, large rooms, vertical structures, steep slopes, and anomalous terrain, we concentrate on the teaming of legged and flying systems that exploit their synergies, while maintaining a high degree of individual autonomy.

The total exploration problem is cast globally and refers to determining which parts of the initially unmapped space $V_{unm}\overset{init}{=} V$ are free $V_{free}\subset V$ or occupied $V_{occ}\subset V$. The environment is represented as an occupancy map $\Ms$ discretizing the volume in (free, occupied or unknown) cubical voxels $m \in \Ms$ with edge size $r_V$. Since most depth sensing modalities cannot pass through objects, the environment may contain hollow, narrow, or more generally occluded sections that can not be explored: 

\begin{definition}[Residual Volume]\label{def:residualSpace}
 Let $\Xi$ be the simply connected set of collision free configurations and $\bar{\mathcal{V}}_m\subseteq \Xi$ the set of all configurations from which the voxel $m$ can be perceived by a depth sensor $\Ss$. Then the residual volume is given as $V_{res} = \bigcup_{m\in \mathcal{M}} ( m \vert\ \bar{\mathcal{V}}_m = \emptyset )$. 
\end{definition}

The multi-robot exploration problem is then defined as: 

\begin{problem}[Multi-Robot Volumetric Exploration]\label{prob:explorationProblem}
 Given a bounded volume $V$ and a team of $N_R$ heterogeneous robots, find a set of $N_R$ collision free paths $\sigma_i$, $i=1,...,N_R$ starting at an initial configuration $\xi_{init}\in \Xi$ that leads to identifying the free and occupied parts $V_{free}$ and $V_{occ}$, such that there does not exist any collision free configuration from which any piece of $V \setminus \{V_{free}, V_{occ}\}$ could be perceived  ($V_{free}\cup V_{occ} = V \setminus V_{res}$). Feasible paths $\sigma_i$ of this problem are subject to the limited Field of View (FoV) of the sensor, its modeled range, and applicable robot motion constraints.
\end{problem}

As noted however, it is in the core principle of this work that multi-robot exploration takes place by robots demonstrating major individual autonomy since subterranean environments present topologies and materials that may prohibit continuous network connectivity. Respectively, we define the single-robot exploration problem given that a volume $V^{S_i^k} \subset V$ is allocated to a robot $i$ at time instance $t_k$ and for a duration $T^i_k$. 

\begin{problem}[Single-Robot Volumetric Exploration]\label{prob:explorationProblem}
 Given a bounded volume $V^{S_i^k}$ and a robot $i$, find a collision free path $\sigma_{S_i^k}$ starting at an initial configuration $\xi_{init,i}^k\in \Xi$ that identifies the free and occupied parts $V_{free}^{S_i^k}$ and $V_{occ}^{S_i^k}$, such that there does not exist any reachable collision free configuration from which any piece of $V^{S_i^k} \setminus \{V_{free}^{S_i^k}, V_{occ}^{S_i^k}\}$ could be perceived. Thus, $V_{free}^{S_i^k}\cup V_{occ}^{S_i^k} = V^{S_i^k} \setminus V_{res}$. A feasible path must respect the sensor model, and applicable motion, namely a max yaw rate and - for legged robots - traversability limitations requiring that $\Ms$ presents supportive ground along every path and its inclination is affordable. 
\end{problem}

%% file: Approach.tex
The presented contribution outlines a strategy for multi-robot ground and flying exploration of predominantly subterranean environments that exploits increased levels of individual robot autonomy based on a comprehensive exploration planner. We detail both the new single-robot Graph-Based exploration Planner (GBPlanner 2.0) and the CoOperation for HeterOgeneous Robot Teams (COHORT) planning method. 

%
\begin{figure*}[h]
\centering
    \includegraphics[width=0.9\textwidth]{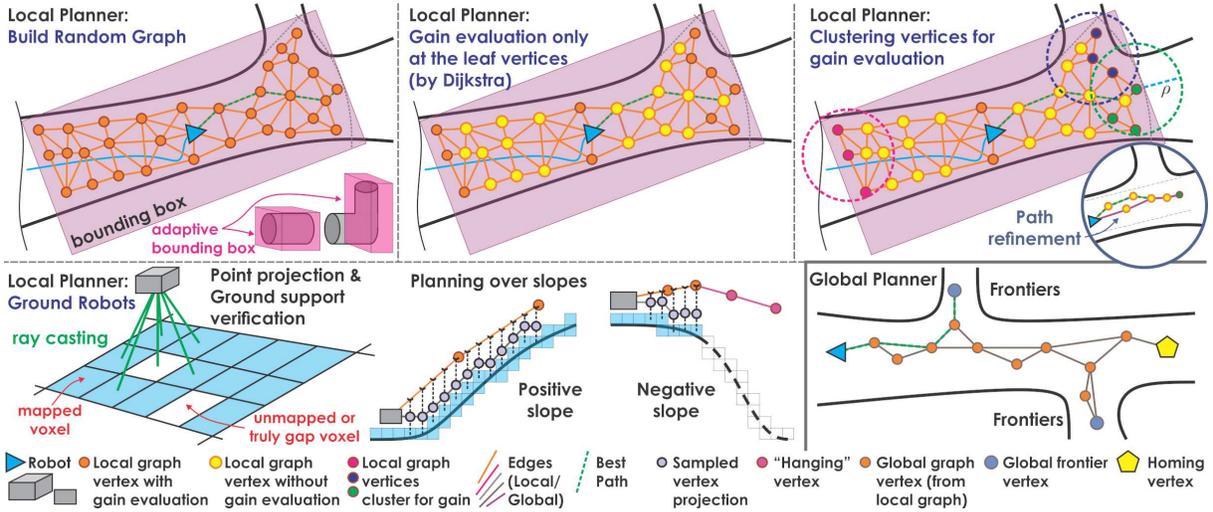}
\vspace{-1ex}
\caption{Outline of the key functional steps of the revised graph-based subterranean exploration planner for aerial and ground (legged) robots.}\label{fig:gbplanner2overview}
\vspace{-3ex}
\end{figure*}
%

\subsection{Graph-based Subterranean Exploration Revisited}

At the core of the presented policy for autonomous subterranean exploration through ground and aerial robot teaming is a path planner for single-robot autonomous exploration of assigned, initially unmapped, underground volumes $V^{S_i^k}$. The method builds on top of our previous open-source work on graph-based subterranean exploration \cite{GBPLANNER_JFR_2020}, which we will refer to as ``GBPlanner1''. The new GBPlanner version presented in this paper, ``GBPlanner2'', has been heavily improved to best handle challenging subterranean geometries and steep slopes, and to provide improved computational performance.
GBPlanner2 retains the bifurcated local/global planning architecture originally proposed, with the local planner being responsible for identifying efficient exploration paths that respect the robot motion and perception constraints, while the global planner is triggered to re-position the robot towards a previously perceived frontier of the exploration space when the local layer reports inability to find a path of significant exploration gain or to ensure that the robot returns home within its endurance limits. Figure~\ref{fig:gbplanner2overview} provides a graphical overview of GBPlanner2. Like GBPlanner1, the new method considers exploration gain, as the new volume to be observed by the robot $i$ if it moves along a certain path $\sigma_{L,best}$ given a depth sensor $\mathbb{S}^i$ with FoV $[F_H^i,F_V^i]$ and modeled range $d_{\max}^i$ (Algorithm~\ref{alg:localgbplanner}).

As compared to GBPlanner1, most changes are in the local exploration step. For ground robots, the method now explicitly handles the case of (steep) positive or negative slopes, vertical obstacles, and ditches, thus offering true $3\textrm{D}$ exploration capabilities for ground systems. Exploiting the volumetric map representation based on Voxblox~\cite{voxblox}, the method randomly samples vertices in the known free space within a local bounding box that is adaptively calculated to best fit the locally explored geometry. To identify the best fitting bounding box $D_L$, Principal Component Analysis (PCA) is performed on the locally aggregated point cloud and the respective normalized eigenvectors $[v_x,v_y,v_z]$ are scaled by $\mu_V>0$ (tunable). Accordingly, a random local graph $\mathbb{G}_L$ can be built. For a flying robot, admissible edge connections between two vertices are simply collision-free straight line segments. Conversely, when planning for a ground robot, the method checks each edge between two vertices by a) projecting points along the edge onto the ground and verifying that ground exists and is mapped, thus being able to support the robot's traversal, and b) verifying that the path inclination is within feasible bounds for the given robot (Algorithm~\ref{alg:buildlocalgraph}). Notably, the projection also queries points on a circle around the shortest point to the ground as ground voxels may occasionally be incorrectly mapped as ``free'' (e.g., due to high LiDAR incidence angle, water puddles). This key feature also allows for the organization of vertices in those that have supportive ground, and those that are in free space but do not have any ground (occupied) voxels below them, called ``hanging'' vertices. They can be connected to other vertices in $\mathbb{G}_L$ through edges adhering to the edge inclination limit and are used for volumetric gain purposes, but their gain is scaled by $e^{-\gamma_H}, \gamma_H>0$ and the edges to them are not allowed to be commanded to the robot. 

\begin{algorithm}[h!] 
\caption{Local Planner}
\label{alg:localgbplanner}
\begin{algorithmic}[1]
\State $\xi_0\gets$ $\mathbf{GetCurrentConfiguration}()$
\State $\mathbb{G}_L\gets\mathbf{BuildLocalGraph}(\xi_0)$ 
\Comment{Algorithm~\ref{alg:buildlocalgraph}}
\State $\Sigma_L \gets \mathbf{GetDijkstraShortestPaths(\mathbb{G}_L, \xi_0)}$ 
\State $\mathbf{ComputeVolumetricGain(\mathbb{G}_L)}$ \Comment{Algorithm~\ref{alg:volumetricGain}}
\State $g_{best}\gets 0$
\State $\sigma_{L,best}\gets \emptyset$
\ForAll{$\sigma \in \Sigma_L$}
 \State $g_{\sigma} \gets \mathbf{ExplorationGain}(\sigma)$ 
 \If{$ g_{\sigma} >g_{best}$}
  \State $g_{best} \gets g_\sigma$; $ \sigma_{L,best} \gets \sigma$ 
 \EndIf
\EndFor
\State $\sigma_{L,best} \gets \mathbf{ImprovePath}(\sigma_{L,best})$
\State \Return $\sigma_{L,best}$
\end{algorithmic}
\end{algorithm}

\begin{algorithm}[h!] 
\caption{Build a Local Graph $\mathbb{G}_L$~$(\xi=[x,y,z,\psi]^T)$}
\label{alg:buildlocalgraph}
\begin{algorithmic}[1]
\Function{BuildLocalGraph}{$\xi_0$} 
\State $\mathbb{V}\gets\{\xi_0\}$; $\mathbb{E}\gets\emptyset$ \Comment{$\xi_0[x_0,y_0,z_0,\psi_0]^T$}
\State $\mathbb{G}_L=(\mathbb{V},\mathbb{E})$
\State $D_L\gets{\mathbf{CalculateAdaptiveLocalBound}(\xi_0)}$ 
\While{$N_\mathbb{V} \leq N_{\mathbb{V},\text{max}}$ \textbf{and} $N_\mathbb{E} \leq N_{\mathbb{E},\text{max}}$}
 \State $\xi_{rand} \gets$ $\mathbf{SampleFree}(D_L)$ 
 \If{GroundRobot}
    \State $\xi_{proj} \gets~\mathbf{ProjectPoint}(\xi_{rand})$
    \State $\xi_{proj}.Hanging \gets~ \textrm{False}$
    \If{not $\mathbf{GroundAttached}(\xi_{proj})$}
        \State $\xi_{proj}.Hanging \gets~\textrm{True}$
    \EndIf
    \State $\xi_{nearest} \gets~ \mathbf{NearestVertex}(\mathbb{G}_L,\xi_{proj})$
    \State \small$e\gets \mathbf{ProjectEdge}(\xi_{proj},\xi_{nearest})$\normalsize
    \State $\xi_{new} \gets~ \xi_{proj}$\normalsize
 \Else
    \State $\xi_{nearest}\gets \mathbf{NearestVertex}(\mathbb{G}_L,\xi_{rand})$
    \State \small$e\gets \mathbf{LineSegment}(\xi_{rand},\xi_{nearest})$\normalsize
    \State $\xi_{rand}.Hanging \gets~ \textrm{False}$
    \State $\xi_{new} \gets~ \xi_{rand}$\normalsize
 \EndIf
 
 \normalsize

 \If{$\mathbf{AdmissibleEdge}(e)$}
  \State $\mathbb{V}\gets \mathbb{V} \cup \{\xi_{new}\}$; $\mathbb{E}\gets \mathbb{E} \cup \{e\}$
  \State $\Xi_{near} \gets \mathbf{NearestVertices}(\mathbb{G}_L,\xi_{new},\delta)$ 
  \ForAll{$\xi_{near} \in \Xi_{near}$}
  \If{GroundRobot}
    \State \small$e_{near} \gets \mathbf{ProjectEdge}(\xi_{new},\xi_{near})$\normalsize
  \Else
    \State \small$e_{near} \gets \mathbf{LineSegment}(\xi_{new},\xi_{near})$\normalsize
  \EndIf
   \If{$\mathbf{AdmissibleEdge}(e_{near})$} 
    \State $\mathbb{E}\gets \mathbb{E} \cup \{e_{near}\}$ 
   \EndIf
  \EndFor
 \EndIf
\EndWhile
\State \Return $\mathbb{G}_L=(\mathbb{V},\mathbb{E})$
\EndFunction
\end{algorithmic}
\end{algorithm}

\begin{algorithm}[h!] 
\caption{Compute Volumetric Gain}
\label{alg:volumetricGain}
\begin{algorithmic}[1]
\Function{ComputeVolumetricGain}{$\mathbb{G}_L$} 
\State $\mathbb{V}_l \gets \mathbf{GetLeafVertices}(\mathbb{G}_L)$
\While{ $\mathbb{V}_l \ne \emptyset$}
    \State $v \gets \mathbb{V}_l[0]$ 
    \State $\mathbb{V}_{near} \gets \mathbf{NearestVertices}(\mathbb{G}_L,v,\rho)$
    \State $v.Gain \gets VolumetricGain(v)$ 
    \If{v.Hanging}
        \State $v.Gain \gets~ e^{-\gamma_H} v.Gain$
    \EndIf
    \ForAll{$v_{near} \in \mathbb{V}_{near}$}
        \State $v_{near}.Gain \gets v.Gain$
        \If{$v_{near}.Hanging$}
            \State $v_{near}.Gain \gets~ e^{-\gamma_H} v_{near}.Gain$
        \EndIf
        \State $\mathbb{V}_{near} \gets \mathbb{V}_{near} \setminus	 v_{near}$
    \EndFor
\EndWhile
\EndFunction
\end{algorithmic}
\end{algorithm}

\vspace{-3ex}
Provided these modifications, the method then builds a local random graph as in GBPlanner1. Once the graph is built, Dijkstra's algorithm is used to derive shortest paths from the robot's location to all vertices. Along these paths $\Sigma_L$, volumetric exploration gain is calculated over each of their vertices and accumulated to derive the total gain of a path as in~\cite{GBPLANNER_JFR_2020}. Since this process is among the most computationally demanding within the method, GBPlanner2 offers the option to calculate the gain only on leaf vertices of the Dijkstra paths and also cluster these vertices using a radius $\rho>0$ thus allowing to approximate the gain of some vertices based on the calculated gain of a nearby vertex. This modification is tailored to computationally-constrained micro flying robots operating in very wide, long and tall environments as ray casting, used to identify the number of unknown voxels to be observed by a depth sensor from a new vertex location, can be very demanding. Algorithms~\ref{alg:localgbplanner}-\ref{alg:volumetricGain} offer an overview of the revised local planning in GBPlanner2. The best path $\sigma_{L,best}$ is derived and conducted by the robot. Beyond utilizing the revised edges from the local layer and other small improvements, the global planning stage of GBPlanner2 is identical to that of GBPlanner1 as in~\cite{GBPLANNER_JFR_2020}.

%
\begin{figure*}[h]
\centering
    \includegraphics[width=0.95\textwidth]{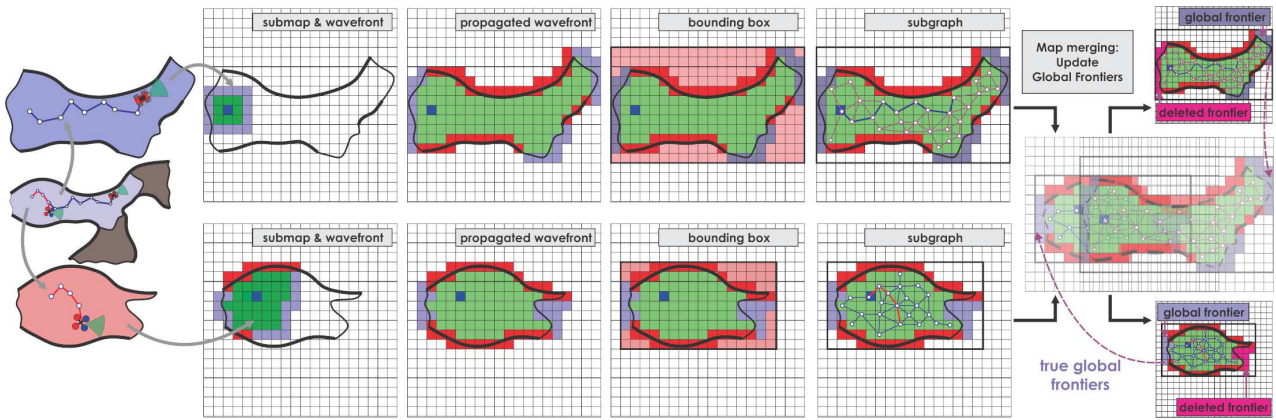}
\vspace{-2ex}
\caption{Outline of how COHORT combines submaps from multiple robots to identify global frontiers and build individual subgraphs, that are merged to form a global graph to allow re-positioning to those frontiers in the global map that are likely to best enable future exploration.}\label{fig:cohortoverview}
\vspace{-3ex}
\end{figure*}
%

\subsection{Cooperation for Heterogeneous Robot Teams}

COHORT enables teamed operation by heterogeneous robotic systems, each empowered with single-agent exploration autonomy based on GBPlanner2. A prerequisite for this task is that all robots' poses and maps are expressed in the same inertial frame $\Is_C$, an ability facilitated by the global multi-modal, multi-robot mapping (M3RM) and Voxgraph frameworks detailed in~\cite{CERBERUS_SUBT_PHASE_I_II,schneider2018maplab,reijgwart2019voxgraph}. M3RM requires software to run both onboard the robot and on a Centralized Computational Hub (C2H), with the robot iteratively sending selective and compressed local map data and the C2H improving and aligning the map estimates of the robots thus requiring a (possibly intermittently available) communication link. COHORT operates as outlined in Figure~\ref{fig:cohortoverview} in order to support the exploration of the unknown volume $V$ (and map $\Ms$) by a team of $N_R$ robots. At any time, every robot $i,~i=1,...,N_R$ builds a local ``submap'' $\Ms_i^l$ which is the volumetric representation of the environment over a predefined window as developed by Voxblox~\cite{voxblox} with resolution $r_C$. The submap window is defined based both on a fixed time-slot (here $45\textrm{ s}$) and a maximum robot displacement from the point of starting a new submap (here $20\textrm{ m}$) with a new submap initiated when any of these thresholds is exceeded.

Once submaps are received and aligned with the global map estimate $\Ms^C$ using Voxgraph~\cite{reijgwart2019voxgraph}, they are processed by COHORT. Upon receiving a new submap, wavefront propagation is performed to identify map frontiers. As wavefront we define the list of unvisited voxels neighboring visited ones in this submap. The wavefront propagation utilizes Breadth-First Search (BFS) and results in getting back adjacent voxels forming common frontiers which are defined as a collection of neighboring frontier points (centroids of unknown voxels neighboring a free mapped voxel). Using the extremities of the visited voxels above, a local bounding box $D_{\Ms_i^l}$ is derived per submap. Within $D_{\Ms_i^l}$ the method randomly samples a fixed number of vertices expressed in the local frame $\Bs_i^l$ of that submap which are also checked for lying in free space. Using the sampled set of free points ($[x,y,z]^T$) in the submap, a graph $\mathbb{G}_{\Ms_i^l}$ (called ``subgraph'') with $N_{C}$ vertices and all collision-free edges within a radius $\rho_C>0$ of each of the sampled points, alongside sampled points over the trajectory of the robot within $\Ms_i^l$ is constructed. Finally, the built $\mathbb{G}_{\Ms_i^l}$ is added to the COHORT's global graph of all robots $\mathbb{G}_C$ which in turn grows to expand through all the explored space from all the robots as subgraphs are integrated into it.

When a new submap is added or the transformation to the submaps changes due to updated multi-robot map alignment, each frontier point is checked in the global map $\Ms^C$ for occupancy. If a frontier point for a submap turns out to lie in a known voxel for another submap, that frontier point is deleted. The frontier points are then clustered into frontiers by identifying all connected components via BFS with 26-connectivity~\cite{rosenfeld1966sequential}. Frontiers that are smaller than a set minimum size $f_{\min}>0$ (here, $f_{\min}=250$) are discarded and their frontier points are deleted. Once each frontier is re-clustered, then the global graph is checked to identify the points of $\mathbb{G}_C$ that are near the geometric centroid of the points forming the frontier. These points are linked to the frontier and facilitate multi-robot exploration coordination. 

Specifically, multi-robot exploration in COHORT takes place by exploiting the above data structure organization and the services of GBPlanner2. In particular, any robot in the team operates by exploring the environment using GBPlanner2 but periodically seeks to update where it explores based on COHORT by re-positioning itself towards a frontier of the map $\Ms^C$ as explored by the team of robots. In detail, as long as a robot is in communication range to the C2H (possibly via multiple network-hops through other robots or breadcrumbed communication nodes as in~\cite{CERBERUS_SUBT_PHASE_I_II}), at every $T_C$ seconds of exploration the following steps are taken: First, the distances from all robots to all points on the COHORT's global graph $\mathbb{G}_C$ and the distances from the home location ($\xi_{init}$) to all $\mathbb{G}_C$ vertices are calculated using Dijkstra. Then, the frontier points are all expressed into the global frame $\Is_C$ and re-clustered such that overlapping frontiers from different submaps are not treated separately and an accurate estimate of the size of each frontier is derived. The re-clustered frontiers are sorted and the best $n_C \%$ are derived. Among the top $n_C \%$ frontiers, the one that is closest to the robot calling for re-positioning is derived and using $\mathbb{G}_C$ the path for the robot to traverse is found, expressed in the robot's coordinate frame, and commanded to the system. COHORT also provides to the robot a new (fixed-size) bounding box $D^{S_i^k}$ updating the GBPlanner2 exploration goal volume $V^{S_i^k}$ such that it is now around the targeted frontier region. Upon arrival to the selected frontier, GBPlanner2 is re-triggered using the new $V^{S_i^k}$ ($D^{S_i^k}$) as goal volume to explore. After $T_C$ seconds of GBPlanner2-based exploration, COHORT will be called again to update where the robot will be exploring, while the procedure continues iteratively on this robot and works in an identical fashion across all robots. Notably, when a robot interacts with COHORT it also stores onboard the location it was at during that interaction. If at a subsequent attempt to interact with COHORT the communication channel is not available, the system will check if the whole assigned $V^{S_i^k}$ is explored and if so, it will return to the point that last offered a communication link (or iteratively proceed to previous such points if needed). If $V^{S_i^k}$ is not fully explored yet, the robot continues for an extra $T_C^e$ seconds using GBPlanner2 before it backtracks to a previous communication point.

%
\begin{figure*}[h]
\centering
    \includegraphics[width=0.99\textwidth]{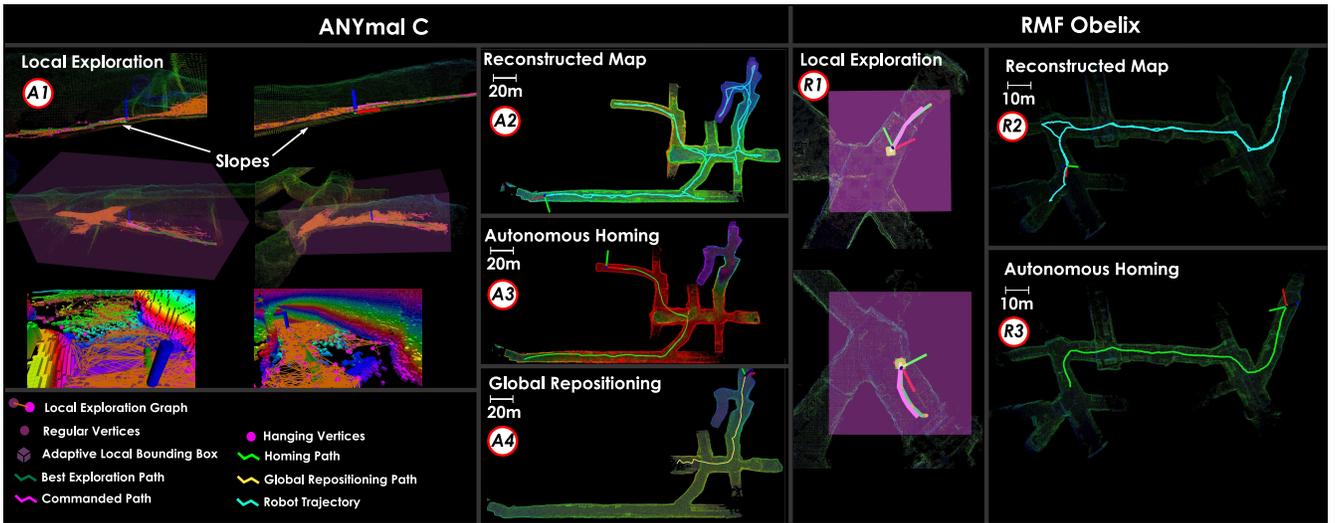}
\vspace{-2ex}
\caption{Autonomous exploration using the ANYmal C legged robot and RMF-Obelix flying system running GBPlanner2 in the Hagerbach underground mine in Switzerland. ANYmal C conducted exploration for $37$ minutes demonstrating navigation over slopes, global repositioning and homing, whereas RMF-Obelix performed an $8$ minute mission including homing. For ANYmal C's mission, $\gamma_H=5$ is used to tune the weight of hanging vertices. }\label{fig:hagerbach}
\vspace{-2ex}
\end{figure*}
%

%% file: EvaluationStudies.tex
To systematically evaluate the proposed contributions, we present two sets of experimental and simulation results. The first involves the field deployment and exploration of different sections of the Hagerbach underground mine in Switzerland using an ANYmal C quadruped robot~\cite{anybotics_website,Miki2022sciro}, and a custom flying robot called ``RMF-Obelix'' (built as a follow-up of the work in~\cite{RMF_ICRA_2021}) with both implementing GBPlanner2 onboard. ANYmal C integrated a Velodyne VLP-16 which was used for the purposes of mapping and volumetric gain evaluation. RMF-Obelix is a lightweight ($1.4~\textrm{kg}$) aerial robot integrating an OUSTER OS0-64 LiDAR. Table~\ref{tab:gbplannerpars} summarizes the GBPlanner2 parameters for ANYmal C and RMF-Obelix, alongside the total robot endurance $T_{end}$, mission time $T_M$, average processing time $t_p$ and mapping resolution $r_V$, while $\mu_V=50~\textrm{m}$. It is noted that both systems rely on our work in~\cite{khattak2020complementary} for localization and mapping, but have very diverse processing capabilities, namely an Intel i7 8850H CPU with 6 cores (12 threads) on ANYmal and an Amlogic A311D ARM big.LITTLE ($4\times$ A73 @2.2 GHz, $2\times$ A53 @1.8 GHz) CPU for RMF-Obelix, and thus have different GBPlanner2 settings. Figure~\ref{fig:hagerbach} presents the results of this mission with ANYmal C exploring for $37~\textrm{min}$ and RMF-Obelix for $8~\textrm{min}$, both without any manual interruptions, automatically re-positioning to different sections of the mine and safely returning to home. We also provide detailed video overview of these missions, available at \url{https://s.ntnu.no/exploration}.

\begin{table}[]
\centering
\vspace{-1ex}
\begin{tabular}{|l|l|l|}
\hline
\textbf{Parameter}          & \textbf{ANYmal C}              & \textbf{RMF-Obelix}            \\ \hline
$r_V$                       & $0.2~\textrm{m}$                & $0.3~\textrm{m}$                \\ \hline
$N_{\mathbb{V},\text{max}}$ & 1000                            & 500                            \\ \hline
$[F_H,F_V],~\textrm{used~}d_{\max}$       & $[360,30]^\circ,~30~\textrm{m}$ & $[360,90]^\circ,~30~\textrm{m}$ \\ \hline
$\rho$                      & $0$ (disabled)                 & $1~\textrm{m}$                  \\ \hline
$T_{end}$                       & $70~\textrm{min}$               & $8.5~\textrm{min}$               \\ \hline
$T_M$                       & $37.6~\textrm{min}$               & $8.2~\textrm{min}$                \\ \hline
$t_p$                       & $605.53~\textrm{ms}$               & $278.59~\textrm{ms}$                \\ \hline
\end{tabular}
\caption{GBPlanner2 parameters for both robots.}
\label{tab:gbplannerpars}
\vspace{-6ex}
\end{table}

%
\begin{figure}[h!]
\centering
    \includegraphics[width=0.99\columnwidth]{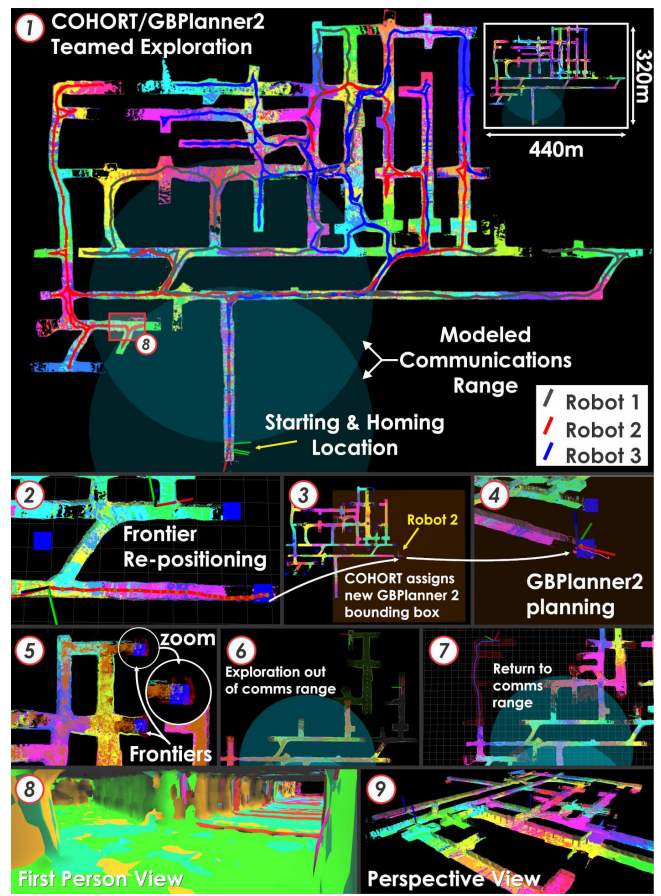} 
\vspace{-4ex}
\caption{Instances of teamed exploration with 3 flying robots showing the combined map at the C2H (1), frontiers (5), and path to frontier outside communication range with new bounds calculated by COHORT (2,3) using its global graph. GBPlanner2 explores outside communication range (shown as cyan circles) and returns after $T_C^e$ (4,6,7). Map $3\textrm{D}$ views in (8,9).}\label{fig:cohortsimulation}
\vspace{-5ex}
\end{figure}
%

To evaluate COHORT and its inter-operation with GBPlanner2, we present a large-scale simulation scenario involving $3$ flying robot models that our team has openly released in the DARPA Subterranean Challenge repositories~\cite{SubtTechRepo} (Robot 1: RMF-Obelix~\cite{RMFSimModel}, Robots 2,3: M100~\cite{M100SimModel}) but with ``virtual endurances'' of $[2000,2500,2500]~\textrm{s}$ respectively. Figure~\ref{fig:cohortsimulation} presents the results for teamed exploration and shows the path of each robot, instances where COHORT re-positioned robots to global frontiers of the commonly explored map (with $r_C=0.4~\textrm{m}$, $N_C=200$, $\rho_C=4.0~\textrm{m}$, $n_C \%=50\%$), and examples of exploration outside of communications range. Every time COHORT commands a robot to re-position itself towards a frontier, GBPlanner2 is engaged ($T_C=50~\textrm{s}$, $D^{S_i^k}=360\times360\times20~\textrm{m}$) onboard the robot. As communication constraints are also modeled, at certain instances GBPlanner2 continues the exploration for an additional $T_C^e=150~\textrm{s}$ before commanding the robot to backtrack to a node where communications were available.

%% file: Conclusion.tex
This paper presented a complete framework for the autonomous teamed exploration of complex environments such as large-scale, multi-branching subterranean settings involving challenging terrain. The method relies on the harmonic synergy between resilient exploration path planning at the level of each individual robot, combined with a multi-robot exploration coordination framework that exploits aligned maps to identify the global frontiers of the mapped environment and is robust against communications-deprived environments. To evaluate our contribution, a set of large-scale field deployments and simulations are presented.